\documentclass{Interspeech2024}

\interspeechcameraready

\title{Improving child speech recognition with augmented child-like speech}

\name[]{Yuanyuan}{Zhang}
\name[]{Zhengjun}{Yue}
\name[]{Tanvina}{Patel}
\name[]{Odette}{Scharenborg}

\address{
  Multimedia Computing Group, Delft University of Technology, the Netherlands}
\email{(y.zhang-44, z.yue, t.b.patel, o.e.scharenborg)@tudelft.nl}


\begin{document}

\renewcommand{\footnotesize}{\scriptsize}

\maketitle

\begin{abstract}
    



State-of-the-art ASRs show suboptimal performance for child speech. The scarcity of child speech limits the development of child speech recognition (CSR). Therefore, we studied child-to-child voice conversion (VC) from existing child speakers in the dataset and additional (new) child speakers via monolingual and cross-lingual (Dutch-to-German) VC, respectively. The results showed that cross-lingual child-to-child VC significantly improved child ASR performance. Experiments on the impact of the quantity of child-to-child cross-lingual VC-generated data on fine-tuning (FT) ASR models gave the best results with two-fold augmentation for our FT-Conformer model and FT-Whisper model which reduced WERs with $\sim$3\% absolute compared to the baseline, and with six-fold augmentation for the model trained from scratch, which improved by an absolute 3.6\% WER. Moreover, using a small amount of “high-quality” VC-generated data achieved similar results to those of our best-FT models.


\end{abstract}

\section{Introduction}

A crucial user group of speech technology is children. In the US alone, for instance, around 12\% of the users of voice assistants are below the age of 12 years \cite{voicestatistics}. 
Automatic speech recognition (ASR) performance of child speech (i.e., child speech recognition; CSR), however, does not achieve the same level as that of adult and adolescent speech \cite{bhardwaj2022automatic, feng2024towards, patel2024child} due to several reasons, including the scarcity of training data, the high pronunciation variability and the high diversity in linguistic phenomena associated with child speech and language development \cite{potamianos2003robust,potamianos1997automatic}. Recently, deep learning-based end-to-end (E2E) ASR models have gained popularity due to the integrated training process and their high performance \cite{Gulati2020conformer, 10301513}. However, the models are data-hungry, which may be problematic for child speech, as typically CSR is a low-resource problem.

To deal with the data scarcity problem, ideally one would want to record (and transcribe) new child speech data in the form of additional speech from existing child speakers in a database or from new child speakers. However, this is often not possible, for various reasons. Therefore, a practical alternative is to artificially generate child(-like) speech. Various augmentation techniques have been investigated for CSR. These include speed perturbation \cite{patel2024child, ko2015audio}, spectral augmentation \cite{singh2022spectral, park2019specaugment, shuyang2023data}, pitch shift \cite{shuyang2023data, chen2020data, zhao23c_interspeech}, and vocal tract length perturbation \cite{singh2022spectral, jaitly2013vocal}. However, most of these techniques primarily modify one or two aspects of the original child or adult speech, e.g. pitch or spectral characteristics, to generate artificial child-like speech, but do not generate new spoken content. Voice conversion (VC) is a good candidate for generating new content: a source speaker's voice is transformed into that of a target speaker while preserving the linguistic content from the source speaker, leading to new content by the target speaker. VC for CSR often involves adult-to-child speech conversion as most of the available speech data is spoken by adult speakers \cite{shahnawazuddin2020voice, singh2021data}. However, due to the differences between adult and child voices, which involve not only pitch but also various other acoustic and perceptual characteristics, the high pronunciation variability and high diversity in linguistic phenomena associated with child speech are lost. Therefore, the resulting speech is child-like rather than child speech. Child-to-child VC, which to the best of our knowledge has not been used before, might solve this problem.

CSR research typically focuses on two scenarios \cite{shuyang2023data, shahnawazuddin2020voice, yu2021slt}: child speech data in the target language is or is not available to (re)train the ASR. In this paper, we aim to improve Dutch CSR performance using VC in these two scenarios. When Dutch child speech is available, we can generate \textit{new spoken content} by the Dutch child speakers in our training set (we refer to this as monolingual VC). However, if no child speech is available in the target language, then monolingual VC into child(-like) speech is not possible because of the absence of target child speakers. To solve this, we propose to use cross-lingual VC \cite{yi2020voice, baas2021voice, zhang-yuanyuan2022_mitigating, zhang2023exploring}, i.e., speech of language A is transformed into child speech of language B. Cross-lingual VC can be applied in both scenarios to create speech from \textit{new child speakers} who produce the same content as the source speakers in our target database. Here, we use German child speech as our target. The cross-lingual VC method is compared to pitch shift, which changes the pitch, and can be viewed as creating a \textit{new speaker} \cite{zhang2023exploring}. For comparison with existing adult-to-child speech results \cite{shahnawazuddin2020voice, singh2021data}, the child-to-child VC method is compared to a teenager-to-child VC method. We use teenager speech for two reasons: 1) our child speech database contains teenager speech but no standard adult speech, 2) ASR performance on teenager speech is fairly close to that of standard adult speech \cite{feng2024towards}.

Additionally, we investigated the effect of the \textit{quantity} and \textit{quality} of the VC-generated speech on CSR performance. For the \textit{quantity} experiments, several folds of VC data were generated. We retrained and fine-tuned our baseline CSR model while gradually adding VC-generated data from each fold. For the \textit{quality} experiments, smaller amounts of VC-generated data with low word error rates (WERs) were used to fine-tune our ASR model and the \textit{Whisper-small} model from OpenAI \cite{radford2023robust}, after which the results were compared.

In summary, this is the first work to explore and compare child-to-child monolingual and cross-lingual VC-based data augmentation, and evaluate the impact of quantity and quality of child-to-child VC speech on CSR performance.

\section{Methodology}
\label{sec:method}
\subsection{Datasets}
\label{subsec:results-disc:datasets}

\textbf{Corpus Gesproken Nederlands (CGN):} The Spoken Dutch Corpus \cite{oostdijk-2000-spoken} contains $\sim$900 hours of Dutch speech recordings from adults aged 18 to 60 years, encompassing various dialects, styles, contexts, and diverse conditions. The speech was recorded both in Flanders (Flemish) and the Netherlands (Dutch). In our experiments, we used the $424.54$ hours of Dutch adult speech to build Dutch ASR baselines.



\noindent\textbf{JASMIN-CGN Corpus:} The JASMIN-CGN corpus \cite{cucchiarini-etal-2006-jasmin} is an extension of the CGN corpus. The corpus includes two speech types: read speech and human-machine interaction (HMI) speech. For our experiments, we used the speech data from Dutch children (age: $6$-$13$ years) and teenagers ($12$-$18$ years). The training data ($7.37$ hours of child speech and $5.32$ hours of teenager speech) denoted by $J_{train}$ was used to train the ASR systems, and to train and infer the VC model. The child speech was only used to train ASR models in the scenario when child speech data is available. The teenager speech was used for both scenarios. 
The test set consists of read (35.73 minutes) and HMI (8.87 minutes) speech from 3 female and 3 male Dutch child speakers, with one child representing each age group from 7-12 years, and was only used for the evaluation of the ASR models. There was no overlap of speakers between the training and test sets.

\noindent
\textbf{KidsTALC Corpus:} KidsTALC \cite{rumberg2022kidstalc} is a corpus containing spontaneous speech recordings by monolingual German children (age: $3.5$-$11$ years) with no or little background noise. The KidsTALC training set consisting of $30$ child speakers (15 males and 15 females) was used to train the VC model and inference to create new child speakers.

\noindent
\textbf{VCTK Corpus:} As additional training data for the VC model, we used the VCTK corpus \cite{vctk}, which is a non-parallel English corpus consisting of speech recordings from $109$ English adult speakers. It contains $\sim$44h of read speech. 


\subsection{Data augmentation}
\label{sub:da}

\subsubsection{VC-based child speech data augmentation }
Our proposed monolingual and cross-lingual child-to-child VC was implemented using AGAIN-VC \cite{chen2021again}, a state-of-the-art (SotA) non-parallel autoencoder-based model. The key steps in the VC data augmentation process are:

\begin{enumerate}

\item \textbf{Training of the VC model}: The AGAIN-VC model is first trained to disentangle speaker and content information from the input speech. 
We followed the training settings in \cite{chen2021again}. The VC model was trained with the VCTK corpus, the Dutch teenager and child speech from $J_{train}$, and German child speech from the KidsTALC corpus to ensure the model's proficiency in converting speech specifically from these demographic groups. The trained VC model was used for all the VC-based data augmentations.

\item \textbf{Source-target speaker pair selection}: 
Theoretically, each source speaker's speech could be converted to any of the $N$ target speakers, leading to an N-fold augmentation of the source speech data. However, such extensive augmentation may not always be necessary or beneficial \cite{bartelds-etal-2023-making}. Therefore, we carefully select the source-target speaker conversion pairs using the approach in \cite{zhang2023exploring}.
To determine these pairs, we extracted compact speaker embeddings using a pre-trained ConvGRU model \cite{spkemb2024} and calculated the cosine similarity between each source-target pair's embeddings \cite{wan2018generalized}. The two most similar target speakers were selected for each source speaker, resulting in a two-fold data augmentation. However, the necessity of this selection method has not been verified in \cite{zhang2023exploring}. To explore the impact of source-target speaker similarity on the augmentation's effectiveness, for each source speaker, in addition to selecting two target speakers with the highest cosine speaker similarity values, two additional methods were studied: selecting two target speakers 1) randomly, and 2) with the lowest cosine speaker similarities.

\item \textbf{Inference of VC models}: Selected source-target pairs were used as input to the trained VC model to generate child speech:


\end{enumerate}

For \noindent\textbf{monolingual VC ($VC_{ml}$)}, both the source and target speakers' speech are in the same language. Specifically, in our $VC_{ml}$ experiments, all the Dutch child speakers from $J_{train}$ were used as the target speakers, to create ``new content''. For the scenario in which child speech data is available, all the Dutch children and teenagers from $J_{train}$ were used as the source speakers to conduct the child-to-child and teenager-to-child experiments, respectively. For the scenario in which child speech data is not available, monolingual VC can not be conducted due to the absence of target child speakers.

For \noindent\textbf{cross-lingual VC ($VC_{cl}$)}, the source and target speakers' speech are in different languages. Specifically, in our $VC_{cl}$ experiments, the German children from KidsTALC were used as the target speakers, to create ``new speakers''. For the scenario in which child speech data is available, all the Dutch children and teenagers from $J_{train}$ were used as the source speakers to conduct the child-to-child and the teenager-to-child experiments, respectively. For the scenario in which child speech data is not available, all the Dutch teenagers from $J_{train}$ were used as source speakers to conduct the teenager-to-child experiment. Compared with monolingual VC, the cross-lingual approach significantly broadens the diversity of child speech in our dataset by incorporating speech patterns and nuances from a different linguistic background. 

\subsubsection{Pitch shift}
We used the pitch shift (PS) method by SoX \cite{sox2021}, in which the pitch of the speech is shifted by “cents”, i.e., 1/100th of a semitone. The resulting pitch makes it sound as if the speech is spoken by a different, new speaker. Therefore, pitch shift increases speaker diversity in the training data \cite{chen2020data}. In our experiments, we generated child-like speech by pitch-shifting the Dutch child and teenager speech. In particular, for each child or teenage speaker, we selected two random shift values between $250$ - $370$ \cite{chen2020data}, resulting in twice a two-fold data augmentation.

\subsection{The ASR models}

All experiments were conducted using the ESPnet toolkit \cite{watanabe2018espnet}. We used a Conformer-based ASR model \cite{Gulati2020conformer} with 12 encoder layers, 5 decoder layers (each with 2048 dimensions), and an attention mechanism with 512 dimensions and 8 attention heads. Training used CTC weight 0.3, attention weight 0.7, and 5,000 BPEs. 
The acoustic features were 80-dim filterbank with 3-dim pitch features. We compared our results to those obtained with a large pre-trained model (Whisper-OpenAI-small \cite{radford2023robust}). All Conformer-based experiments were conducted on four NVIDIA GeForce GTX 1080 Ti GPUs and all Whisper-based experiments were on one 1080 Ti GPU.

We compared two training methods: When training from scratch the Conformer models were trained for 35 epochs and each epoch had 10k iterations. The final model was averaged over the top 10 models with the lowest training loss \cite{huang2016snapshot}. For the fine-tuning experiments, both the Conformer and Whisper-small models were fine-tuned for 15 epochs of 500 iterations. The final model was averaged over the top 5 models with the lowest training loss.

\subsection{Experiments}
\label{sub:exp-setups}

\subsubsection{Data augmentation experiments}
\label{subsub:caseI/II}

\noindent\textbf{Without child speech training data:} 
Two baseline models were trained only on Dutch adult speech (from CGN) and on Dutch adult+teenager (from JASMIN-CGN) speech to tease apart the influence of database mismatch between training and test set (from JASMIN-CGN). The \textit{Base1} model training set (adult+teenager) was used for the following experiments. Subsequently, two experiments were conducted with the generated child(-like) speech from: 1) the Dutch teenager-to-German child $VC_{cl}$; 2) pitch-shifted teenager speech.



\noindent\textbf{With child speech training data:} 
The second baseline (\textit{Base2}) model was trained using the Dutch adult, teenager, and child speech. Subsequently, six experiments were carried out with the generated child(-like) speech from: 1) pitch-shifted child speech; 2) the Dutch child-to-German child $VC_{cl}$; 3) the Dutch child-to-Dutch child $VC_{ml}$; 4) pitch-shifted teenager speech 5) the Dutch teenager-to-German child $VC_{cl}$; 6) the Dutch teenager-to-Dutch child $VC_{ml}$.

\subsubsection{Quantity and quality experiments}
\label{subsub:quantity}
To investigate the effect of the amount of VC-generated artificial speech on CSR performance, incremental sets of child-to-child speech by cross-lingual VC were generated. Additionally, we expanded the dataset by augmenting the number of child speech utterances to two, four, ..., and ten folds by selecting two, four, ..., and ten target speakers with the highest speaker similarity values for each source speaker. These sets were incrementally added to the training set of \textit{Base2} to retrain the Conformer-based ASR models. Moreover, the quantity experiments were also conducted by fine-tuning both the \textit{Base2} and Whisper \cite{radford2023robust} models using the incrementally added VC-generated speech and original Dutch child speech. This was to study the robustness of the proposed child-to-child $VC_{cl}$ method to different models.


To investigate the effect of the quality of the VC-generated speech on CSR performance, we selected high-quality artificial speech by filtering the VC-generated speech through the \textit{Base2} model, selecting only those utterances that exhibited the lowest WERs. This process creates data sets containing utterances with progressively higher WER thresholds, from 10\%, 20\%, and so on, up to the inclusion of all speech data. The quality experiments were then conducted by fine-tuning both the \textit{Base2} and Whisper models using the filtered VC-generated speech and original Dutch child speech. 

\section{Results and Discussion}
\subsection{Data augmentation experiments:}
We first evaluated the \textit{Base0} model on the CGN adult speech test sets \cite{van2009results}. The model achieved WERs of ~8.1\% for broadcast news (BN) and 24.8\% for conversational telephone speech (CTS), which is comparable to SotA results using a language model (unlike this study; BN: 6.6\%, CTS: 21.6\%)  \cite{feng2024towards}.

\noindent\textbf{Without child speech training data}:  
Table 1 shows the performance of the two baseline models (trained only on adult speech and on adult+teenager speech) and the three augmented models tested on Dutch child speech. As expected, \textit{Base0} (trained on adult speech) exhibited high WERs when tested on child speech, likely due to a database mismatch (CGN vs. JASMIN-CGN) and the differences between adult and child speech. Adding teenager speech to the training set resulted in an average reduction of WER by 15\% absolute. 
For read speech, adding the pitch-shifted teenager speech improved CSR performance significantly\footnote{Matched Pairs Sentence-Segment Word Error (MAPSSWE) \cite{gillick1989some} was carried out as in \cite{test2024} to perform statistical significance tests between baselines (\textit{Base1 or Base2}) and augmented models. We reported p-values.}. However, no improvement was found when adding the $VC_{cl}$-generated speech from Dutch teenagers, potentially due to the fact that the target speech used to create $VC_{cl}$ was spontaneous. 
For HMI speech, both adding the pitch-shifted teenager speech and the $VC_{cl}$-generated speech from Dutch teenagers significantly improved CSR performance.




\begin{table}[t]
\centering
\label{Tab:CaseIandII}
\caption{WERs of data augmentation experiments on child test sets. Teen is teenagers. Bold indicates the lowest WER for each test set in each of the two scenarios. The blue row indicates the lowest WERs in all the data augmentation experiments.} 
\resizebox{0.9\columnwidth}{!}{%

\begin{tabular}{l|c|c|c|c}
\toprule
\multicolumn{5}{c}{\textbf{Without child speech training data}}\\
\midrule
&&\multicolumn{3}{c}{\textbf{WER\%}}\\
\cmidrule{3-5}
\textbf{Training data}   &\textbf{hours}   & \textbf{Read} & \textbf{HMI} & \textbf{Avg.}    \\ \midrule
Base0: adult                      &  $424.5$  & $46.8$     & $48.9$    & $47.2$   \\ 
\cmidrule{2-5}
Base1: adult + teen              &  $429.9$    & $32.3$     & $34.4$    & $32.7$ \\
\cmidrule{2-5}
Base1 + teen (PS)  & $440.5$ &  $\textbf{26.7}^{\ast\ast\ast}$  &$\textbf{23.4}^{\ast\ast\ast}$   &  $\textbf{26.0}$  \\ 
Base1 + teen ($VC_{cl}$) &  $438.9$ & $32.2$     & $24.8^{\ast\ast\ast}$    & $30.8$  \\
\toprule
\midrule

\multicolumn{5}{c}{\textbf{With child speech training data}}\\
\toprule
Base2: Base1 + child         &    $437.2$          & $10.5$	     & $20.3$  & 	$12.3$  \\
\cmidrule{2-5}
Base2 + child (PS)&  $452.0$ & $9.0^{\ast\ast}$	  & $19.0$     & $10.8$  \\ 
\rowcolor{lightcornflowerblue}
Base2 + child ($VC_{cl}$)& $449.3$& $\textbf{7.4}^{\ast\ast\ast}$   & 	$\textbf{16.6}^{\ast\ast}$  & 	$\textbf{9.1}$     \\ 
Base2 + child ($VC_{ml}$) & $449.3$  &   $7.7^{\ast\ast\ast}$	  &  $18.6$	   & $9.7$  \\
\cmidrule{2-5}
Base2 + teen (PS)        &    $447.9$         &$9.5^{\ast}$	    & $19.1$	    & $11.3$   \\ 
Base2 + teen ($VC_{cl}$) & $446.3$   & $11.4$   & 	$17.8^{\ast}$	    & $12.6$   \\ 
Base2 + teen ($VC_{ml}$) & $446.3$   &   $9.7$  &   	$19.3$  &	$11.5$  \\


\toprule

\end{tabular}%
}
\\* $p < .05$; ** $p < .001$; *** $p< .001$.
\vspace{-4mm}
\end{table}

\noindent\textbf{With child speech training data}: 
Table 2 shows the performance of the \textit{Base2} model (trained on adult+teenager+child speech) and the six augmented models tested on Dutch child speech. As expected, adding child speech to the training data resulted in an average reduction of WER by $\sim$20\% absolute compared to the \textit{Base1} model. For read speech, adding the pitch-shifted child speech, and cross-lingual and monolingual VC-generated speech from child speech individually improved the CSR performance significantly. However, when using teenager speech, only for pitch-shifted data a significant improvement was found, while adding $VC_{ml}$ and $VC_{cl}$ speech from teenagers did not change performance. This lack of improvement is likely due to the lack of child speech linguistic phenomena in the source teenager speech, resulting in speech that is less child-like.
For HMI speech, although both pitch-shifted child speech and $VC_{ml}$-generated speech from child speech showed an improvement compared to \textit{Base2}, this improvement was not significant. Adding $VC_{cl}$-generated speech from child speech, however, did improve performance significantly. This same pattern of results was observed when using teenager-based augmented speech.
 The best performance was obtained with data generated through child-to-child $VC_{cl}$ with an average reduction of WER by 3.2\% absolute.

In summary, when no child speech data is available, adding teenager pitch-shifted data leads to significant improvements over our strong baseline. These results are in line with \cite{chen2020data, shuyang2023data}, and extend their results on Mandarin and Icelandic to Dutch. For the more spontaneous HMI speech, adding cross-lingual VC using teenager speech also led to a significant performance improvement. Adding more spontaneous speech to the training data was beneficial, even though the speech data was not child speech. This is in line with recent findings that adding more diverse training data, even if not of the target data, improves recognition performance \cite{zhang2023exploring}. When child speech is available for ASR training, adding any child-speech-based augmentation gave significant improvements over baseline, which overall outperformed the results obtained when using teenager speech, with the VC methods outperforming pitch shift. These results show that child-to-child VC outperforms adult-to-child VC \cite{shahnawazuddin2020voice, singh2021data}. The best results are obtained with cross-lingual VC, which (slightly) outperformed monolingual child-to-child VC, suggesting that diversifying the speakers in the training data outperforms adding more content from existing speakers.

\begin{table}[t]
\label{SimilarityTable}
\centering
\vspace{-5mm}
\caption{Top/Random/Last VC source-target pairs selections with the scores and corresponding WERs on child test sets.}
\resizebox{0.73\columnwidth}{!}{%
\begin{tabular}{c|c|c|c|cl}
\toprule
\textbf{}   &\textbf{Avgerge Cosine} & \multicolumn{3}{c}{\textbf{WER\%}}  \\
\cmidrule{3-5}
\textbf{Selection}   &\textbf{Similarity} & \textbf{Read} & \textbf{HMI} & \textbf{Avg.}   \\
\midrule
\rowcolor{lightcornflowerblue}
Top & 0.59 & \textbf{7.4}   & 	\textbf{16.6}  & 	\textbf{9.1}     \\ 
Random &0.46 &   7.4  & 		17.3	 & 	9.2    \\ 
Last & 0.29  & 8.2   & 		18.9  & 		10.2    \\ 
\bottomrule
\end{tabular}
}
\vspace{-4mm}
\end{table}

\noindent\textbf{The influence of speaker similarity for  $VC_{cl}$ data augmentation.} 
In Table 2, ``Top'' indicates the highlighted line in Table 1. Using two \textit{Random} speakers gave the same result for read speech, despite a lower cosine similarity score, but led to a small WER drop for HMI speech. Using the two speakers with the lowest cosine similarity score (\textit{Last}) reduced performance for both speech types. Therefore, source-target speaker matching is essential for optimal VC-generated speech for improving CSR performance.



\subsection{Quantity and quality experimental results}

\noindent\textbf{Quantity}: The left panels of Figure 1 show the effect of the data quantity for the three models for read (top) and HMI speech (middle), and averaged over both speaking styles (bottom). For read speech, comparing the three models, all models showed a big drop in WER when the VC-generated data was increased two-fold, with a WER of 7.9\% for Base2-FT, 6.8\% for Whisper-FT, and 7.4\% for training from scratch. Adding more data only led to a further improvement for the model trained from scratch, with the best performance for the six-fold data augmentation (WER of 6.3\%). For HMI speech, the picture is rather different: although adding two-fold augmented data led to a large improvement for the model trained from scratch, overall, adding more data led to performance degradation. This degradation might stem from spontaneous speech recognition is a more difficult task than read speech recognition \cite{feng2024towards}, requiring high-quality augmented speech. Overall, both FT models achieved their best results using two-fold augmented data, where the \textit{Whisper-FT} model (blue line, 9.1\% WER) slightly outperformed the \textit{Base2-FT} model (orange line, 9.6\% WER). Despite the Whisper model's initially high WERs (read: 43.3\%, HMI: 55.3\%).

\begin{figure}[t]
\vspace{-3mm}
    \centering
    \label{QuantityExpFigure}
    \includegraphics[width=\columnwidth]{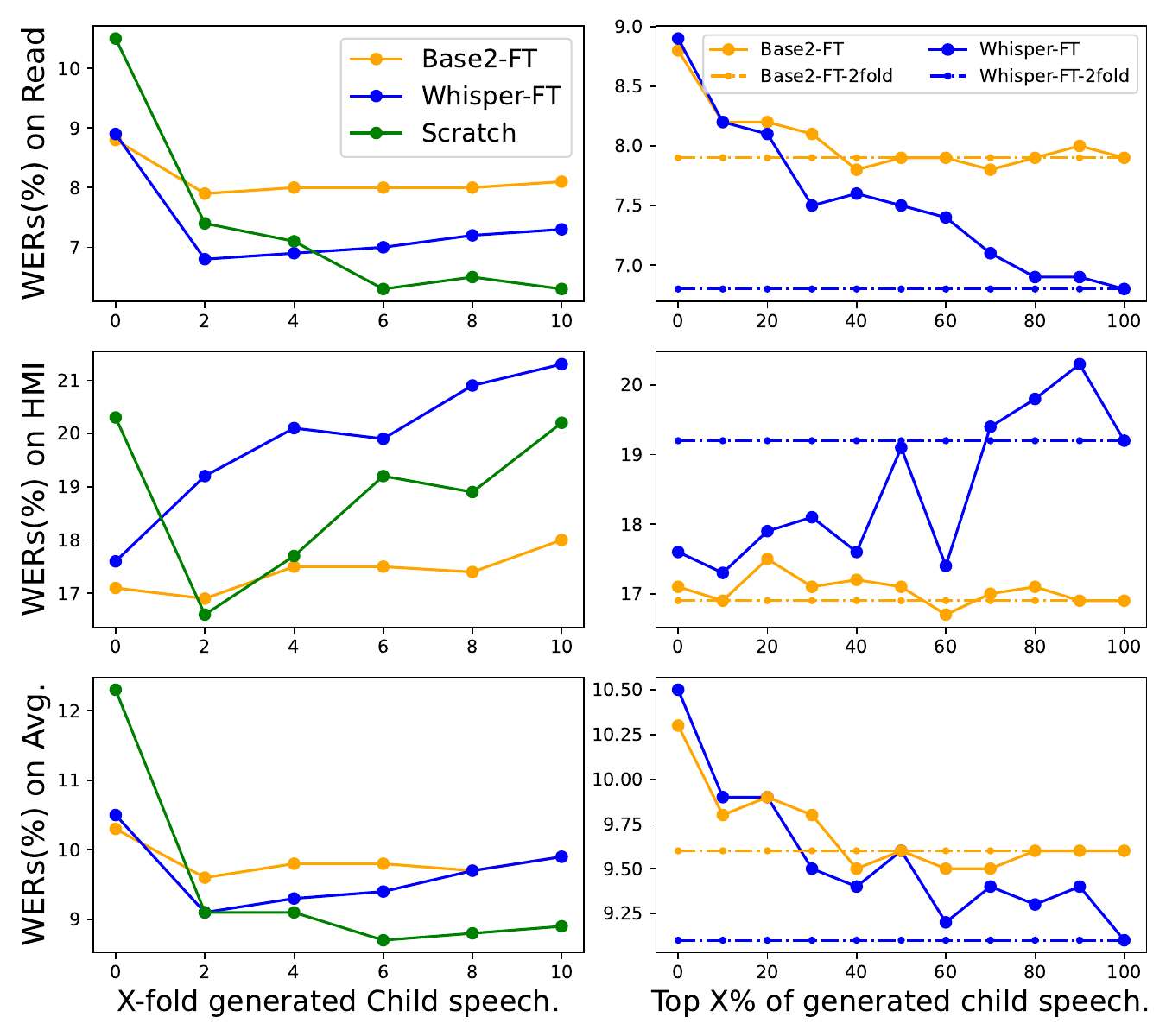}  
    \caption{(Left panels) Quantity experiments with the \textit{Base2} model trained from scratch (green), after fine-tuning (orange) and Whisper with fine-tuning (blue) on child + x-fold child speech generated by  $VC_{cl}$. (Right panels) Quality experiments: Conformer \textit{Base2} and Whisper with FT on child speech + lowest 10\%, 20\%,...,90\% WER speakers of the two-fold $VC_{cl}$ data. Dashed lines: results of both models on the two-fold $VC_{cl}$ data.} 
    \label{fig:exp pipline}
    \vspace{-5.5mm}
\end{figure}



\noindent\textbf{Quality}: 
The right panels of Figure 1 show the results of the data quality experiments for fine-tuning the two models for read (top) and HMI speech (middle), and averaged over both speaking styles (bottom). For read speech, adding increasingly more data, even of less quality, continued to improve recognition performance for Whisper FT, with the best result for the full set. For \textit{Base2}-FT, the best result was obtained when adding the top 40\% best recognized data. Therefore, a smaller training set of higher quality outperformed a larger training set. 
For HMI speech, adding more data decreased performance; the best WER was found when only adding 10\% of the data with lower WERs. 
Performance was rather stable across all data set sizes for the \textit{Base2}-FT model. 

Previous research, e.g., \cite{shahnawazuddin2020voice, singh2021data}, typically used a fixed amount of all converted child-like speech, our experiments showed that two-fold $VC_{cl}$ augmented speech in some cases is sufficient to achieve good CSR performance. Therefore, increasing the amount of generated data is not always beneficial \cite{baas2021voice, bartelds-etal-2023-making}. Our quality results show adding less, higher quality child speech by $VC_{cl}$ for fine-tuning leads to similar or even improved performance for both the Whisper and Conformer ASR models compared to adding more data.





\section{Conclusion}
In this study, we investigated improving CSR performance using child-to-child VC-generated speech (monolingual and cross-lingual) when child speech was available for training an ASR. Compared with the baseline CSR model, child-to-child VC-generated speech outperformed teenager-to-child VC-generated speech; cross-lingual child-to-child VC-generated speech outperformed pitch-shifted Dutch child speech. When no child speech data was available, augmenting pitch-shifted speech from teenagers gives the best results.
The experiments on the quantity of cross-lingual VC-generated speech data showed that augmenting the speech dataset two-fold with child-to-child cross-lingual VC-generated speech is sufficient to achieve good CSR performance, both when training models from scratch and in fine-tuning scenarios. 
Adding a small amount of `high-quality' VC-generated speech achieved performance levels comparable to our best fine-tuned models. 
\bibliographystyle{IEEEtran}
\bibliography{mybib}

\begin{thebibliography}{10}
\providecommand{\url}[1]{#1}
\csname url@samestyle\endcsname
\providecommand{\newblock}{\relax}
\providecommand{\bibinfo}[2]{#2}
\providecommand{\BIBentrySTDinterwordspacing}{\spaceskip=0pt\relax}
\providecommand{\BIBentryALTinterwordstretchfactor}{4}
\providecommand{\BIBentryALTinterwordspacing}{\spaceskip=\fontdimen2\font plus
\BIBentryALTinterwordstretchfactor\fontdimen3\font minus \fontdimen4\font\relax}
\providecommand{\BIBforeignlanguage}[2]{{%
\expandafter\ifx\csname l@#1\endcsname\relax
\typeout{** WARNING: IEEEtran.bst: No hyphenation pattern has been}%
\typeout{** loaded for the language `#1'. Using the pattern for}%
\typeout{** the default language instead.}%
\else
\language=\csname l@#1\endcsname
\fi
#2}}
\providecommand{\BIBdecl}{\relax}
\BIBdecl

\bibitem{voicestatistics}
R.~Paul, ``Voice search statistics 2024 (usage $\&$ emerging trends),'' [online]. Available: https://www.answeriq.com/voice-search-statistics/.

\bibitem{bhardwaj2022automatic}
V.~Bhardwaj \emph{et~al.}, ``Automatic speech recognition {(ASR)} systems for children: A systematic literature review,'' \emph{Applied Sciences}, vol.~12, no.~9, p. 4419, 2022.

\bibitem{feng2024towards}
S.~Feng, B.~M. Halpern, O.~Kudina, and O.~Scharenborg, ``Towards inclusive automatic speech recognition,'' \emph{Computer Speech \& Language}, vol.~84, p. 101567, 2024.

\bibitem{patel2024child}
\BIBentryALTinterwordspacing
T.~Patel and O.~Scharenborg, ``Improving end-to-end models for children’s speech recognition,'' \emph{Applied Sciences}, vol.~14, no.~6, 2024. [Online]. Available: \url{https://www.mdpi.com/2076-3417/14/6/2353}
\BIBentrySTDinterwordspacing

\bibitem{potamianos2003robust}
A.~Potamianos and S.~Narayanan, ``Robust recognition of children's speech,'' \emph{IEEE Transactions on speech and audio processing}, vol.~11, no.~6, pp. 603--616, 2003.

\bibitem{potamianos1997automatic}
A.~Potamianos, S.~Narayanan, and S.~Lee, ``Automatic speech recognition for children,'' in \emph{Proc. of Eurospeech}, 1997, pp. 2371--2374.

\bibitem{Gulati2020conformer}
A.~Gulati, J.~Qin, C.~Chiu, N.~Parmar, Y.~Zhang, J.~Yu, W.~Han, S.~Wang, Z.~Zhang, Y.~Wu, and R.~Pang, ``Conformer: Convolution-augmented transformer for speech recognition,'' in \emph{Proc. INTERSPEECH}, 2020, pp. 5036--5040.

\bibitem{10301513}
R.~Prabhavalkar, T.~Hori, T.~N. Sainath, R.~Schlüter, and S.~Watanabe, ``End-to-end speech recognition: A survey,'' \emph{IEEE/ACM Transactions on Audio, Speech, and Language Processing}, vol.~32, pp. 325--351, 2024.

\bibitem{ko2015audio}
T.~Ko, V.~Peddinti, D.~Povey, and S.~Khudanpur, ``Audio augmentation for speech recognition,'' in \emph{Proc. INTERSPEECH}, 2015, pp. 3586--3589.

\bibitem{singh2022spectral}
V.~Singh, H.~Sailor, S.~Bhattacharya, and A.~Pandey, ``Spectral modification based data augmentation for improving end-to-end asr for children's speech,'' \emph{arXiv preprint arXiv:2203.06600}, 2022.

\bibitem{park2019specaugment}
D.~S.~Park, W.~Chan, Y.~Zhang, C.~Chiu, B.~Zoph, E.~D.~Cubuk, and Q.~V.~Le, ``{SpecAugment: A Simple Data Augmentation Method for Automatic Speech Recognition},'' in \emph{Proc. INTERSPEECH}, 2019, pp. 2613--2617.

\bibitem{shuyang2023data}
Z.~Shuyang, M.~Singh, A.~Woubie, and R.~Karhila, ``Data augmentation for children asr and child-adult speaker classification using voice conversion methods,'' in \emph{Proc. INTERSPEECH}, 2023.

\bibitem{chen2020data}
G.~Chen, X.~Na, Y.~Wang, Z.~Yan, J.~Zhang, S.~Ma, and Y.~Wang, ``Data augmentation for children's speech recognition--the" ethiopian" system for the {SLT} 2021 children speech recognition challenge,'' \emph{arXiv preprint arXiv:2011.04547}, 2020.

\bibitem{zhao23c_interspeech}
S.~Zhao, M.~Singh, A.~Woubie, and R.~Karhila, ``{Data augmentation for children ASR and child-adult speaker classification using voice conversion methods},'' in \emph{Proc. INTERSPEECH}, 2023, pp. 4593--4597.

\bibitem{jaitly2013vocal}
N.~Jaitly and G.~Hinton, ``Vocal tract length perturbation (vtlp) improves speech recognition,'' in \emph{Proc. ICML Workshop on Deep Learning for Audio, Speech and Language}, vol. 117, 2013, p.~21.

\bibitem{shahnawazuddin2020voice}
S.~Shahnawazuddin, N.~Adiga, K.~Kumar, A.~Poddar, and W.~Ahmad, ``Voice conversion based data augmentation to improve children's speech recognition in limited data scenario.'' in \emph{Proc. INTERSPEECH}, 2020, pp. 4382--4386.

\bibitem{singh2021data}
D.~Singh, P.~P. Amin, H.~Sailor, and H.~A. Patil, ``Data augmentation using cyclegan for end-to-end children asr,'' in \emph{Euro. Sig. Process. Conf. (EUSIPCO)}, 2021, pp. 511--515.

\bibitem{yu2021slt}
F.~Yu, Z.~Yao, X.~Wang, K.~An, L.~Xie, Z.~Ou, B.~Liu, X.~Li, and G.~Miao, ``The slt 2021 children speech recognition challenge: Open datasets, rules and baselines,'' in \emph{Spoken Language Technology Workshop (SLT)}.\hskip 1em plus 0.5em minus 0.4em\relax IEEE, 2021, pp. 1117--1123.

\bibitem{yi2020voice}
Z.~Yi, W.~Huang, X.~Tian, J.~Yamagishi, R.~Das, T.~Kinnunen, Z.~Ling, and T.~Toda, ``Voice conversion challenge 2020—intra-lingual semi-parallel and cross-lingual voice conversion,'' in \emph{Proc. Joint Workshop for the Blizzard Challenge and Voice Conversion Challenge}, 2020, pp. 80--98.

\bibitem{baas2021voice}
M.~Baas and H.~Kamper, ``Voice conversion can improve asr in very low-resource settings,'' in \emph{Proc. INTERSPEECH}, 2022, pp. 3513--3517.

\bibitem{zhang-yuanyuan2022_mitigating}
Y.~Zhang, Y.~Zhang, B.~M. Halpern, T.~Patel, and O.~Scharenborg, ``{Mitigating bias against non-native accents},'' in \emph{Proc. INTERSPEECH}, 2022, pp. 3168--3172.

\bibitem{zhang2023exploring}
Y.~Zhang, A.~Herygers, T.~Patel, Z.~Yue, and O.~Scharenborg, ``Exploring data augmentation in bias mitigation against non-native-accented speech,'' in \emph{Automatic Speech Recognition and Understanding Workshop (ASRU)}.\hskip 1em plus 0.5em minus 0.4em\relax IEEE, 2023, pp. 1--8.

\bibitem{radford2023robust}
A.~Radford, J.~Kim, T.~Xu, G.~Brockman, C.~McLeavey, and I.~Sutskever, ``Robust speech recognition via large-scale weak supervision,'' in \emph{Int. Conf. on Machine Learning}.\hskip 1em plus 0.5em minus 0.4em\relax PMLR, 2023, pp. 28\,492--28\,518.

\bibitem{oostdijk-2000-spoken}
N.~Oostdijk, ``The {Spoken Dutch Corpus}. {O}verview and first evaluation,'' in \emph{{Proc. Int. Conf. on Lang. Resources and Eval. ({LREC})}}, Athens, Greece, May 2000.

\bibitem{cucchiarini-etal-2006-jasmin}
C.~Cucchiarini, H.~V. hamme, O.~van Herwijnen, and F.~Smits, ``{JASMIN}-{CGN}: Extension of the {Spoken Dutch Corpus} with speech of elderly people, children and non-natives in the human-machine interaction modality,'' in \emph{{Proc. Int Conf on Lang. Resources and Eval. ({LREC})}}, Genoa, Italy, May 2006.

\bibitem{rumberg2022kidstalc}
L.~Rumberg \emph{et~al.}, ``{kidsTALC}: A corpus of 3-to 11-year-old german children’s connected natural speech,'' in \emph{Proc. INTERSPEECH}, 2022.

\bibitem{vctk}
\BIBentryALTinterwordspacing
Y.~Junichi, V.~Christophe, and M.~Kirsten, ``{CSTR VCTK} corpus: English multi-speaker corpus for {CSTR} voice cloning toolkit (v. 0.92),'' Univ. of Edinburgh. The Centre for Speech Technology Research (CSTR), 2019. [Online]. Available: \url{https://doi.org/10.7488/ds/2645}
\BIBentrySTDinterwordspacing

\bibitem{chen2021again}
Y.~Chen, D.~Wu, T.~Wu, and H.~Lee, ``{Again-vc}: A one-shot voice conversion using activation guidance and adaptive instance normalization,'' in \emph{Proc. of IEEE Int. Conf. on Acous., Speech and Sig. Process. (ICASSP)}, 2021, pp. 5954--5958.

\bibitem{bartelds-etal-2023-making}
M.~Bartelds, N.~San, B.~McDonnell, D.~Jurafsky, and M.~Wieling, ``Making more of little data: Improving low-resource automatic speech recognition using data augmentation,'' in \emph{Proceedings of the 61st Annual Meeting of the Association for Computational Linguistics (Volume 1: Long Papers)}, A.~Rogers, J.~Boyd-Graber, and N.~Okazaki, Eds.\hskip 1em plus 0.5em minus 0.4em\relax Association for Computational Linguistics, Jul. 2023, pp. 715--729.

\bibitem{spkemb2024}
``Simple speaker embeddings.'' [online]. Available: https://github.com/RF5/simple-speaker-embedding.

\bibitem{wan2018generalized}
L.~Wan, Q.~Wang, A.~Papir, and I.~Moreno, ``Generalized end-to-end loss for speaker verification,'' in \emph{Proc. of IEEE Int. Conf. on Acous., Speech and Sig. Process. (ICASSP)}, 2018, pp. 4879--4883.

\bibitem{sox2021}
B.~Chris, R.~Mans, Robs, and K.~Ulrich, ``{S}o{X} - {S}ound e{X}change,'' [online]. Available: http://sox.sourceforge.net.

\bibitem{watanabe2018espnet}
S.~Watanabe \emph{et~al.}, ``{ESPnet: End-to-End Speech Processing Toolkit},'' in \emph{Proc. INTERSPEECH}, 2018, pp. 2207--2211.

\bibitem{huang2016snapshot}
G.~Huang, Y.~Li, G.~Pleiss, Z.~Liu, J.~E. Hopcroft, and K.~Q. Weinberger, ``Snapshot ensembles: Train 1, get m for free,'' in \emph{International Conference on Learning Representations}, 2016.

\bibitem{van2009results}
D.~A. van Leeuwen, J.~Kessens, E.~Sanders, and H.~Heuvel, ``Results of the n-best 2008 dutch speech recognition evaluation,'' in \emph{Proc. INTERSPEECH}, 2009.

\bibitem{gillick1989some}
L.~Gillick and S.~J. Cox, ``Some statistical issues in the comparison of speech recognition algorithms,'' in \emph{Proc. of IEEE Int. Conf. on Acous., Speech and Sig. Process. (ICASSP)}.\hskip 1em plus 0.5em minus 0.4em\relax IEEE, 1989, pp. 532--535.

\bibitem{test2024}
``{WER} statistical significance test.'' [online]. Available: https://github.com/talhanai/wer-sigtest.

\end{thebibliography}

\end{document}